\documentclass[11pt]{article}
\usepackage[margin=0.8in]{geometry} 
\usepackage[english]{babel}
\usepackage{soul}
\usepackage{xcolor}
\usepackage{tabulary,booktabs,multirow}
\usepackage{listings}
\usepackage{hyperref}
\usepackage[font=small,labelfont=bf,skip=16pt]{caption}
\usepackage{subcaption}
\usepackage{tikz,float}
\usepackage[ruled,vlined]{algorithm2e}
\usepackage{orcidlink}
\usetikzlibrary{bayesnet,arrows}
\tikzset{>=latex}
\usepackage{amsmath,amssymb,bm}
\usepackage{algorithm2e} 
\usepackage{cleveref}
\usepackage{cuted}
\usepackage[printonlyused,withpage]{acronym}

\DeclareMathOperator{\E}{\mathbb{E}}
\DeclareMathOperator{\R}{\mathbb{R}}

\DeclareMathOperator*{\argmax}{arg\,max}

\DeclareMathAlphabet{\mathcal}{OMS}{cmsy}{m}{n}

\usepackage{ntheorem}
\theoremseparator{:}

\lstdefinestyle{pythonstyle}{
    language=Python,
    basicstyle=\ttfamily\small,
    breaklines=false,
    commentstyle=\color{green!50!black},
    keywordstyle=\color{blue},
    stringstyle=\color{red},
    numbers=left,
    numberstyle=\tiny\color{gray},
    numbersep=5pt,
    frame=single,
    framesep=15pt,
    rulecolor=\color{black!30},
    showstringspaces=false,
    tabsize=4
}
\lstdefinestyle{console} {
	language=bash,
	basicstyle=\footnotesize\ttfamily\singlespacing,
	backgroundcolor=\color[RGB]{248, 247, 242},
	upquote=true,
	frame=single,
	framerule=0pt,
	breakindent=0pt,
	breaklines=true,
	aboveskip=0pt,
	belowskip=4pt,
	xleftmargin=10pt,
	xrightmargin=10pt
}

\title{LaGarNet: Goal-Conditioned Recurrent State-Space Models for Pick-and-Place Garment Flattening}

\author{
Halid Abdulrahim Kadi$^{1}$, Student Member,~IEEE~\orcidlink{0000-0001-9290-467X} and 
Kasim Terzi{\'c}$^{1}$, Member,~IEEE~\orcidlink{0000-0001-6692-209X}
\thanks{$^{1}$~School of Computer Science, University of St Andrews.}
}

\begin{document}

\maketitle

\begin{abstract}
We present a novel goal-conditioned recurrent state space (GC-RSSM) model capable of learning latent dynamics of pick-and-place garment manipulation. Our proposed method LaGarNet matches the state-of-the-art performance of mesh-based methods, marking the first successful application of state-space models on complex garments. LaGarNet trains on a coverage-alignment reward and a dataset collected through a general procedure supported by a random policy and a diffusion policy learned from few human demonstrations; it  substantially reduces the inductive biases introduced in the previous similar methods. We demonstrate that a single-policy LaGarNet achieves flattening on four different types of garments in both real-world and simulation settings. 
\end{abstract}

\section{Introduction} \label{sec:introduction}

Achieving garment flattening is a critical milestone for accomplishing the difficult laundry-folding task \cite{doumanoglou2016folding}. A learning-based control method can effectively perform tasks by learning and utilising a world model of environments \cite{ha2018world, hafner2019learning,hafner2020dream, deng2023facing}. However, such world models have been extremely rare in the cloth-manipulation domain due to the difficulty of capturing accurate deformation and addressing self-occlusion of cloth objects \cite{ma2021learning,yan2021learning,lin2020softgym}. 

\begin{figure}[!ht]
    \centering
    \subfloat[\centering Synthetic observation of garments and square workspace in simulation. \label{fig:sim-garments}] {{\includegraphics[width=0.25\textwidth]{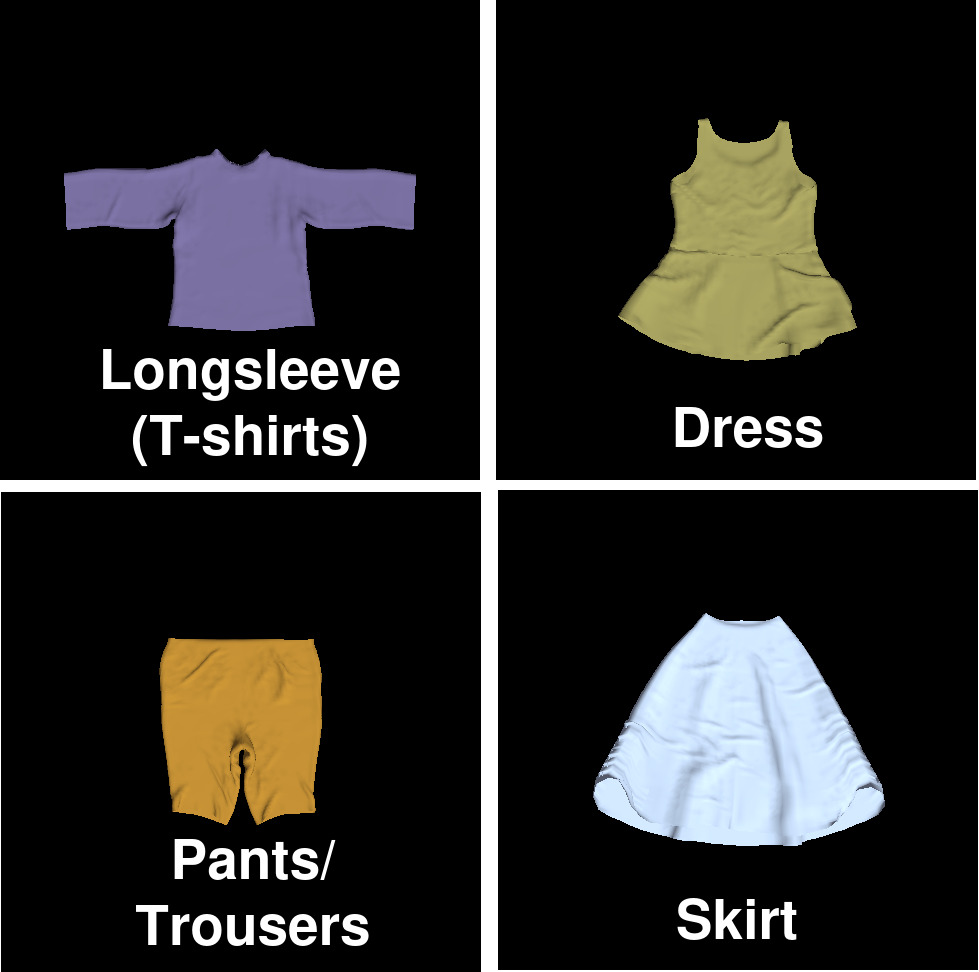} }}\;
     \subfloat[\centering \textit{UR3e} real-world observation and ring-shape workspace \label{fig:ur3e-ring}] {{\includegraphics[width=0.45\textwidth]{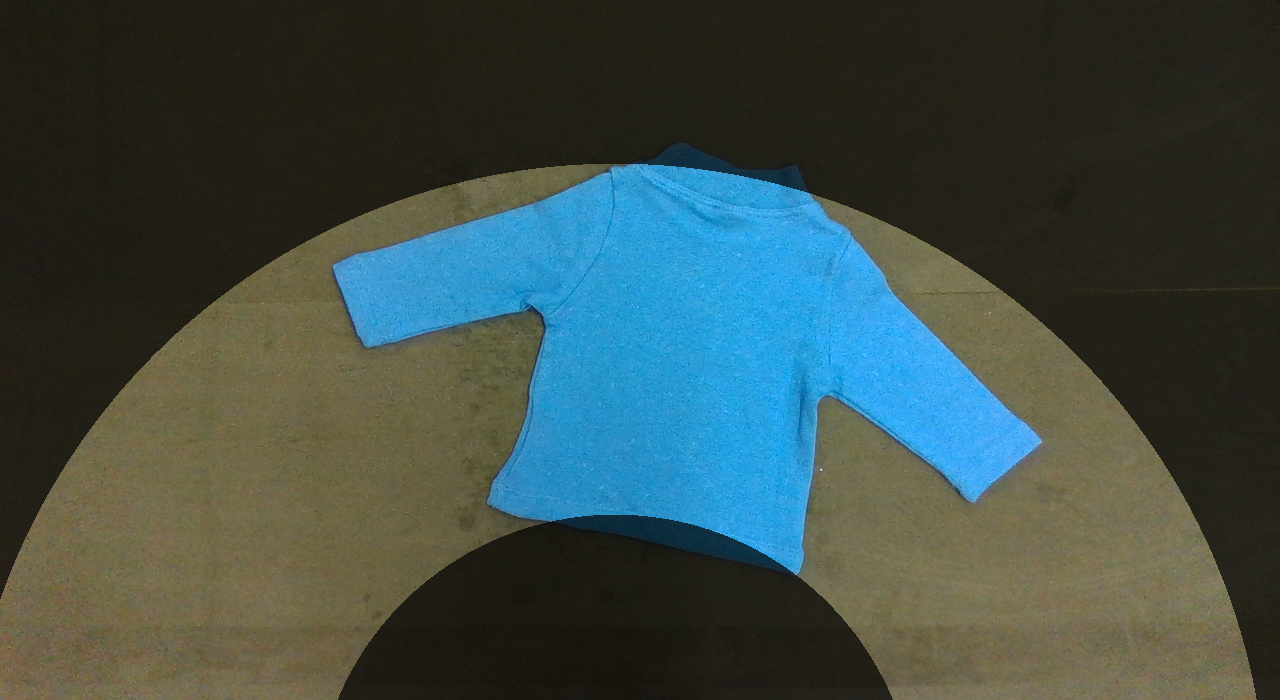} }}
    \caption{Goals of garment-flattening environments. The synthetic garment data is from \textit{Cloth3D} dataset \cite{bertiche2020cloth3d} simulated in \textit{SoftGym} \cite{lin2020softgym}. We train our \textit{LaGarNet} policies on the dataset collected in simulation environments \cite{canberk2023cloth} and transfer them to \textit{UR3e} real-world environments using \textit{DRAPER} deployment framework \cite{kadi2025draper}.   \label{fig:clothfunnels-garments}} \label{fig:garments-goals}
\end{figure}

\begin{figure}[!ht]
    \centering
    \includegraphics[width=0.6\textwidth]{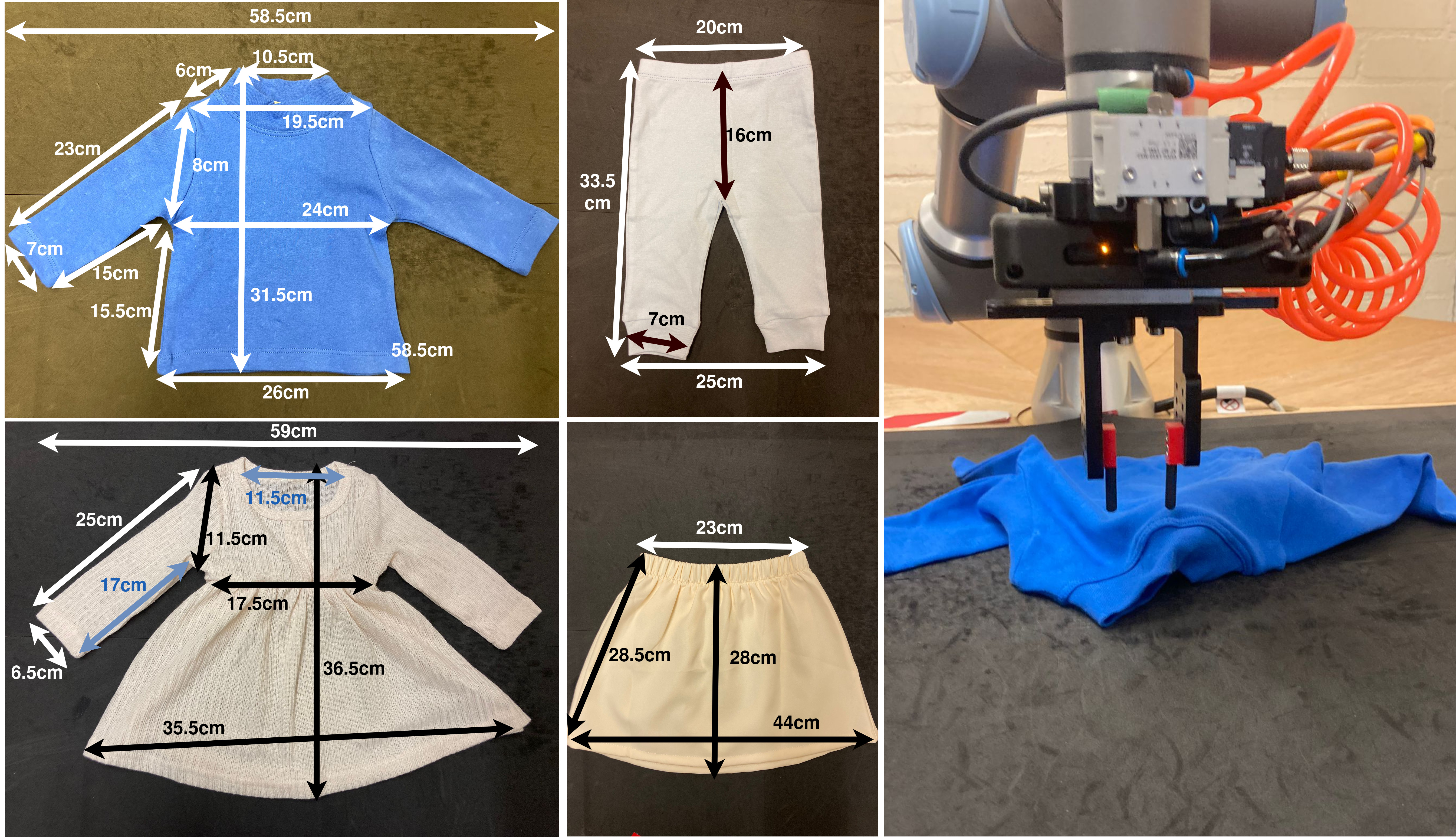}
    \caption{Real garments used for real-world experiments with a \textit{UR3e} robot arm.  The longsleeved T-shirt consists of 95\% cotton and 5\% elastane, with a royal blue colour. The dress is composed of 100\% cotton and features a beige colour. The trousers are made of 100\% cotton and are presented in light blue. The skirt is crafted from 100\% polyester in a khaki shade. }
     \label{fig:real-garments}
\end{figure}

Recurrent State-Space Model (RSSM) \cite{hafner2019learning}, a popular instance of world models, was first successfully applied by the domain-specific latent dynamic planning algorithm \textit{PlaNet-ClothPick} \cite{kadi2024planet}. This algorithm can perform towel-flattening tasks in the real world \cite{kadi2025draper}, but its RSSM is reliant on a specially engineered reward function as well as on a dataset collected through a particular blend of expert and random oracle policies in simulation. This makes it difficult to generalise to more complex garment shapes because it requires engineering individual expert oracles for each type of garment, many of which are complex. This paper presents an RSSM-based model capable of performing complex garment flattening tasks (see Figure \ref{fig:garments-goals}), mitigates strong inductive biases, and generalises the capacity of world models to a wider range of cloth-manipulation tasks. We achieve this by proposing a goal-conditioned RSSM (GC-RSSM) model trained on a general data collection procedure supported by an expert diffusion policy as well as a general coverage-alignment reward function for cloth flattening.

Most successful garment-flattening systems use dual robot arms with hybrid action primitives, including \text{pick-and-place}, \text{pick-and-fling} and \text{pick-and-drag}. Although single-gripper pick-and-place (\text{PnP}) primitives require significantly more iterations and longer execution times compared to dual-arm manipulation \cite{avigal2022speedfolding, ha2022flingbot, canberk2023cloth}, our human-in-the-loop experiments demonstrate that it can achieve excellent flattening and folding outcomes across different types of garments. This gap between human and robot performance on garment flattening under the same observation-action constraints highlights an interesting and challenging machine learning problem to investigate the application of world models in garment manipulation. 



Furthermore, robot arms are often fixed on a table in cloth manipulation robotic research \cite{lin2022learning,lee2024learning,canberk2023cloth}, leading to a ring-shaped workspace for PnP primitives as shown in Figure \ref{fig:ur3e-ring}. Real garments are generally larger than simple fabrics, and it becomes physically impossible to flatten the garments on a fixed square window within the workspace of a single robot arm, which is common practice in this domain \cite{kadi2025draper, hoque2022visuospatial, lee2021learning}. Thereby, it requires us to develop a transfer strategy to bridge the gap between observation and the workspace, and the gap between the simulation and the real robot environment.

\begin{figure*}[t!]
    \centering
    \includegraphics[width=1.0\textwidth]{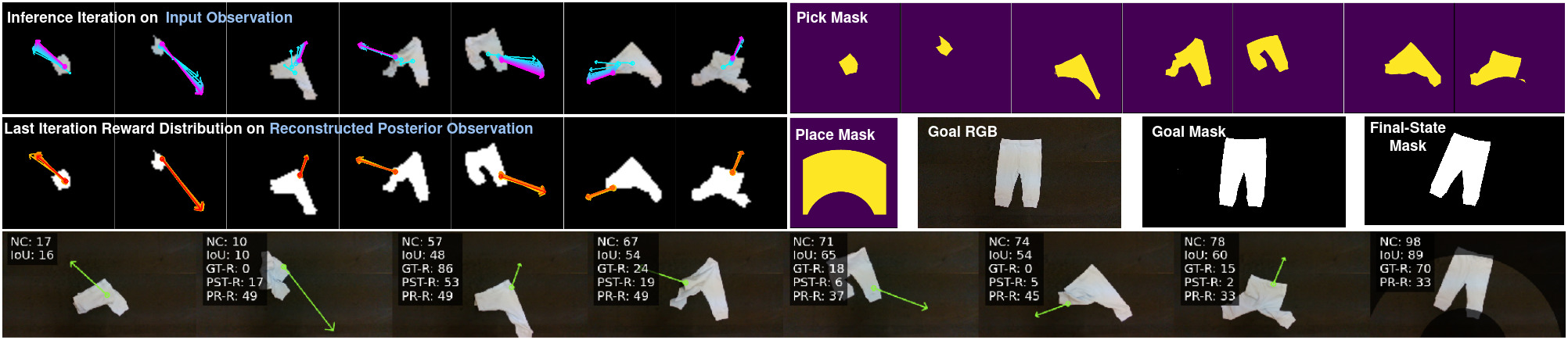}
    \caption{A qualitative demonstration of \textit{LaGarNet} flattening real trousers. In the top right row, bright blue arrows represent the mean of the action Gaussian distribution in the early iteration stages, while the purple arrow represents the later resulting stages. In the row bellow, yellow arrows represent the sampled actions with predicted rewards close to 0, and red arrows represent those close to 0.5. The bottom row demonstrates overall flattening effect of \textit{LaGarNet} in a real-robot setup with normalised coverage (\text{NC}), maximum intersection-over-union (IoU), ground-truth reward (GT-R), predicted posterior reward (PST-R) and predicted prior reward (PR-R); all values are multiplied by 100. }
     \label{fig:ur3e-real-action-inference}
\end{figure*}

In general, we propose a general garment-flattening model-based deep reinforcement learning method, \textit{LaGarNet}, that can flatten four different types of complex garments, utilising the whole workspace of a robot arm in the real world (Figure \ref{fig:ur3e-real-action-inference}). \textit{LaGarNet} mitigates the strong inductive biases presented in its previous counterpart  \textit{PlaNet-ClothPick} \cite{kadi2024planet}. Our single-policy \textit{LaGarNet}  consistently achieves above 94\% normalised coverage (\text{NC}), 87\% normalised improvement (\text{NI}) and 79\% maximum intersection-over-union \text{(Max IoU)} between the current and goal cloth observations across all examined garment types in simulation. It reaches on average 87\% NC, 77\% NI and 79\% Max IoU in our real-world experiments on real examined garments as shown in Figure \ref{fig:real-garments}. Our mesh-free latent planning method LaGarNet matches the reported state-of-the-art performance of mesh-based planning methods on corresponding garment flattening \cite{lin2022learning,huang2022mesh}. In a nutshell, we make the following four contributions:

1. We propose a goal-conditioned recurrent state-space model (GC-RSSM) and develop a single goal-conditioned garment-flattening planning method \textit{LaGarNet} that can flatten all four examined types of garments in both simulation and real-world environments;

2. We suggest a general data-collection strategy for offline training of \text{PnP} garment-flattening controllers, which is a mixture of an expert \textit{Diffusion Policy} learned from few human demonstrations and a masked-biased random policy;

3. We design a novel and effective coverage-alignment reward function for general garment flattening that can be calculated both in simulation and real-world environments;

4. We develop a workspace sampling and transfer heuristic to deploy the trained policy from a fixed square window observation space to a constrained environment limited by the  ring-shape workspace of robot arms for \text{PnP} primitives.

\section{Related Work} \label{sec:related-workd}

\noindent Applications of learning-based methods on cloth-shaping tasks, i.e., flattening and folding cloth-like objects, include model-free reinforcement learning \cite{matas2018sim,wu2019learning,ha2022flingbot,hietala2021closing,lin2020softgym,hietala2021closing,lee2021learning, canberk2023cloth, he2024fabricfolding, gu2023learning, avigal2022speedfolding}, \text{model-based reinforcement learning} \cite{yan2021learning,hoque2022visuospatial,ma2021learning,arnold2021cloth,lin2020softgym,lin2022learning, huang2022mesh} and  \text{behaviour cloning} \cite{lips2024learning,weng2022fabricflownet, teng2022multidimensional, mo2022foldsformer, huang2024sis, chi2023diffusionpolicy} algorithms. Many successful learning-based  controllers employ intermediate explicit representations to assist downstream controllers to mitigate the server self-occlusion problem of cloth. These encompass key-point detection \cite{lips2024learning, gusseme2025insights}, seam feature extraction \cite{huang2024sis}, mesh reconstruction \cite{huang2022mesh} as well as optical flow \cite{weng2022fabricflownet} and dense visual correspondence between current and goal images \cite{wu2024unigarmentmanip}. In contrast, our work focuses on learning a latent dynamic representation and generating actions through planning, making our method generalisable to different scenarios.

These systems also vary in terms of action space. Whist few methods attempt low-level position/velocity control \cite{matas2018sim,lin2020softgym, hietala2021closing,ha2022flingbot}, some adopts  \text{pick-and-fling} (PnF) action primitives \cite{doumanoglou2014autonomous}, \text{pick-and-place} (PnP)  \cite{sun2015accurate}, and \text{pick-and-drag} (PnD) primitives \cite{he2024fabricfolding} to achieve cloth shaping. Some  approaches based hybrid primitives  of the aforementioned three action primitives have demonstrated significant efficacy in facilitating diverse garment manipulation tasks \cite{avigal2022speedfolding, canberk2023cloth}. However, the incorporation of these hybrid strategies introduces additional complexity to the action space for data collection and control itself. Single-gripper PnP only suffers from misgrasping, multi-layer grasping as well as cloth slipping issues \cite{lin2022learning, hoque2022visuospatial}; it has been investigated and addressed from both hardware and control perspectives \cite{kadi2025draper, hoque2021lazydagger}. In this paper, we focus on PnP primitive and put the focus on the learning aspects of the latent dynamic model.

\subsection{Model-Based Methods in Cloth Shaping} \label{sec:mbrl}

The use of cloth dynamics is prevalent in reinforcement learning approaches; learning and using such dynamics can improve data efficiency of policy training and improve the precision of actions during inference time \cite{hafner2019learning, hafner2023mastering, kadi2024planet, lin2022learning}. It is also effective for goal-conditioned tasks \cite{duan2024learning, hoque2022visuospatial}, as we can exploit the distance between current and goal states for planning. The model-based methods in this domain can be classified into mesh-based and mesh-free methods, depending if the cloth-mesh reconstructed during test time.

Most model-based methods focus on reconstructing a 3D mesh and mesh dynamic of the target cloth \cite{huang2022mesh, lin2022learning, tian2025diffusion}. For example, \textit{Visible Connectivity Dynamics (VCD)} \cite{lin2022learning} first applies such mesh models  on the visible part of the cloth to achieve more precise planning. Building on \textit{VCD}, Huang et al.~\cite{huang2022mesh} also adopt point clouds as input for mesh reconstruction and planning in the mesh-dynamic space; their method \textit{MEDOR}  explicitly reconstructs the occluded part of the cloth from a top-down camera by fine-tuning an initial reconstructed mesh in real time based on the template of the target garment. 

On the other hand, the mesh-free model-based approach \textit{Contrastive Forward Modelling (CFM)} \cite{yan2021learning} trains latent dynamics for planning applications. Hoque et al.~\cite{hoque2022visuospatial} propose \textit{Visual Foresight Modelling (VSF)}, which integrates variational video prediction models within the \textit{Visual MPC} framework \cite{ebert2018visual} to address model exploitation. Nevertheless, \textit{VSF} demonstrates limited efficacy in garment-flattening applications \cite{lin2022learning} and neither of these mesh-free planning methods have  managed to match the performance of mesh-based \textit{VCD} \cite{lin2022learning}.  In contrast, our proposed mesh-free planning method \textit{LaGarNet} matches the reported state-of-the-art performance the above mesh-based methods on flattening tasks while having a smaller numbers of parameters and faster inference time for each action step.

As a mesh-free method, \textit{Deep Planning Network (PlaNet)} \cite{hafner2019learning} has been widely studied in control tasks \cite{yan2021learning, ma2021learning, lin2020softgym}, but shows limited efficacy in deformable object manipulation. This is attributed to blurred visual reconstructions that lose edge and corner details critical for shaping \cite{yan2021learning, hoque2022visuospatial, lin2022learning}. \textit{PlaNet-ClothPick} \cite{kadi2024planet} mitigates this by restricting pick actions to the cloth mask, designing a flattening-specific reward, and using an engineered offline dataset. While it achieves perfect flattening in simulation and over 70\% NC on real towels \cite{kadi2024planet, kadi2025draper}, it fails to generalise to complex garments due to strong inductive biases. Our single-policy \textit{LaGarNet} reduces these biases through a general data-collection procedure and coverage-alignment reward, enabling robust garment flattening.

Lastly, model-based methods leverage the difference between current and goal observations to generate reward signals for cloth shaping \cite{hoque2022visuospatial, yan2021learning}. Arnold et al.~\cite{arnold2021cloth} extend this idea to mesh-level goal-conditioned policies for complex tasks, but their method remains limited to fabric manipulation. Model-free approaches train goal-conditioned representations \cite{mo2022foldsformer}, flow networks \cite{weng2022fabricflownet}, or use \textit{Hindsight Replay Buffer} \cite{andrychowicz2017hindsight, lee2021learning}, yet are restricted to simple fabrics. In contrast, we propose a goal-conditioned RSSM (GC-RSSM) that predicts step-wise flattening rewards and achieves garment flattening across four types: long-sleeved T-shirts, trousers, skirts, and dresses.

\subsection{Reward Functions in Cloth Shaping} \label{sec:reward}
Reward engineering is an essential component for the success of a reinforcement learning (RL) controller. Many RL algorithms in garment-flattening systems come with their own reward function \cite{canberk2023cloth, gu2023learning, avigal2022speedfolding}. \textit{ClothFunnels} \cite{canberk2023cloth} adopts a canonicalisation-alignment reward  function $\mathcal{R}_{NA}$  which is a linear interpolation of normalised delta deformation distance for alignment $\mathcal{R}_{dA}$ and normalised delta rigid distance for canonicalisation $\mathcal{R}_{dN}$, i.e., $\mathcal{R}_{NA} = \alpha\mathcal{R}_{dN} + (1-\alpha)\mathcal{R}_{dA}$. \textit{Learning2Unfold} \cite{gu2023learning} employs a coverage-orientation reward $\mathcal{R}_{CO}$ that has the similar interpolation between the normalised coverage $\mathcal{R}_C$ and the orientation $\mathcal{R}_o = 1 - tanh^2\theta$, where $\theta = \alpha|\theta_P| + (1-\alpha)|\theta_F|$; the $\theta_P$ represents the angle between the current cloth and goal cloth and $\theta_F$ represent the flipping angle $\theta_F$ to assure the cloth is facing down.

\textit{SpeedFolding} \cite{avigal2022speedfolding} uses a coverage-smoothness reward function $\mathcal{R}_{CS}$ that mixes the delta coverage $\mathcal{R}_{dC} = NC_t - NC_{t-1}$ and delta smoothness estimation $\Tilde{R}_{dS}$ between the current and the last frame, then clip reward between 0 and 1 after a tanh scaling, i.e., $\mathcal{R}_{CS} = max(tanh(\alpha*\mathcal{R}_{dC} + \beta*\Tilde{R}_{dS}), 0)$; note that the smoothness estimation is inferred through a pre-trained smoothness prediction network. \textit{PlaNet-ClothPick} employs a specially engineered reward function based on \textit{VSF}'s reward specifically for the fabric-flattening task \cite{hoque2022visuospatial}. In this study, we propose a new reward function coverage-alignment $\mathcal{R}_{CA}$ that simplifies the \textit{SpeedFolding}'s coverage-smoothness reward $\mathcal{R}_{CS}$ \cite{avigal2022speedfolding}. Note that, unlike \textit{ClothFunnels}'s canonicalisation-alignment reward which needs to calculate particle-wise distances, which requires an oracle, our reward function should, in theory, work for both simulation and real-world learning.


\subsection{Data Collection and Preparation}

The high-dimensional state-action space of deformable objects makes it challenging to train policy representations, dynamics models, or value functions. Moreover, model-based methods in cloth-shaping tasks often suffer from model exploitation (choosing actions the model incorrectly estimates as optimal) and compounding error (accumulated errors in imagined roll-outs). The effectiveness of a learned world model depends on how diverse and comprehensive the trajectory data are in the training dataset \cite{duan2024learning}. Training typically requires a large amount of data collected with carefully designed blends of policies \cite{hoque2022visuospatial, yan2021learning, ma2021learning, kadi2024planet}. To address this, prior work has used demonstration data \cite{matas2018sim,hoque2022visuospatial}, corner-biased sampling \cite{hoque2022visuospatial,weng2022fabricflownet} (biasing pick actions toward fabric corners and edges), and other engineered strategies \cite{arnold2021cloth,lin2022learning}. However, these approaches impose strong inductive biases on the learning algorithm. In contrast, our study proposes a general data collection procedure based on a diffusion policy trained from a few human demonstrations and a random oracle policy, which empirically yields effective training data for MBRL algorithms.

\section{LaGarNet} \label{sec:planet-clothpick++}

The proposed \textit{LaGarNet} method for garment flattening is an offline model-based deep reinforcement learning approach. We achieve generalisation on garment flattening through training a goal-conditioned recurrent state-space model (Section \ref{sec:gc-rssm}) on a more unified data collection procedure in simulation (Section \ref{sec:data-collection}) and a coverage-alignment reward function designed for all garment-flattening tasks (Section \ref{sec:ca-reward}). 




\subsection{Goal-Conditioned Recurrent State Space Model} \label{sec:gc-rssm}

State-space models (SSMs) are mainly used as generative frameworks for sequential data, supporting tasks such as inferring future states and learning representations \cite{ma2024interpretable}. Recurrent State-Space Model (RSSM) \cite{hafner2019learning}, using a recurrent neural network, is defined under the Partially Observable Markov Decision Process (POMDP) formulation with the following latent state dynamic: (1) recurrent dynamic model $\bm{h}_t = f(\bm{h}_{t-1}, \bm{z}_{t-1}, \bm{a}_{t-1})$, (2) representation model $\hat{\bm{z}}_t \sim q(\hat{\bm{z}} \; | \; \bm{h}_t, \bm{x}_t)$, and (3) transition predictor $\tilde{\bm{z}}_t \sim p(\tilde{\bm{z}} \;|\; \bm{h}_{t})$, where $\bm{x}$ represents observation, $\bm{a}$ represents action, $\bm{h}$ represents the deterministic latent representation, and $\hat{\bm{z}}$ and $\tilde{\bm{z}}$ represent the prior and posterior stochastic latent states. The latent dynamic model of the original \textit{PlaNet-ClothPick} is developed under this formulation.

\begin{figure}[t]
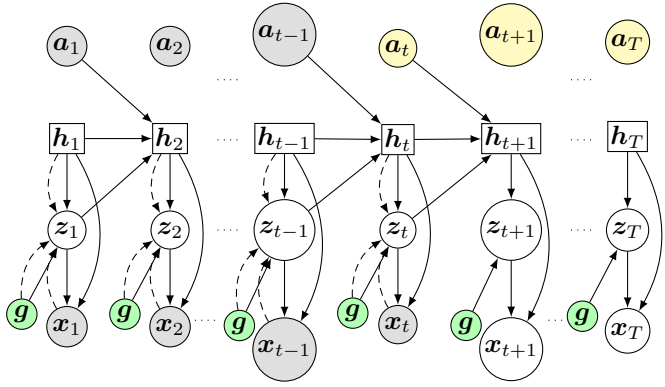

\centering
  \tikz{
 \node[latent, minimum size=0.4cm] (z1) {$\boldsymbol{z}_1$};
 \node[latent, minimum size=0.4cm, right=of z1, xshift=-0.15cm] (z2) {$\boldsymbol{z}_2$};
 \node[latent, minimum size=0.4cm, right=of z2, xshift=-0.15cm] (zt) {$\boldsymbol{z}_{t-1}$};
 \node[latent, minimum size=0.4cm, right=of zt, xshift=-0.15cm] (zt1) {$\boldsymbol{z}_{t}$};
 \node[latent, minimum size=0.4cm, right=of zt1, xshift=-0.15cm] (zt2) {$\boldsymbol{z}_{t+1}$};
 \node[latent, minimum size=0.4cm, right=of zt2, xshift=-0.15cm] (zT) {$\boldsymbol{z}_T$};

 \node[latent,  rectangle , minimum size=0.4cm, above=of z1, yshift=-0.25cm] (h1) {$\boldsymbol{h}_{1}$};%
 \node[latent,  rectangle , minimum size=0.4cm, above=of z2, yshift=-0.25cm] (h2) {$\boldsymbol{h}_{2}$};%
 \node[latent,  rectangle , minimum size=0.4cm, above=of zt, yshift=-0.4cm] (ht) {$\boldsymbol{h}_{t-1}$};%
 \node[latent,  rectangle , minimum size=0.4cm, above=of zt1, yshift=-0.25cm] (ht1) {$\boldsymbol{h}_{t}$};%
 \node[latent,  rectangle , minimum size=0.4cm, above=of zt2, yshift=-0.4cm] (ht2) {$\boldsymbol{h}_{t+1}$};%
 \node[latent,  rectangle , minimum size=0.4cm, above=of zT, yshift=-0.25cm] (hT) {$\boldsymbol{h}_{T}$};%

 \node[obs, minimum size=0.4cm, above=of h1, yshift=-0.25cm] (a1) {$\boldsymbol{a}_1$}; %
 \node[obs, minimum size=0.4cm, above=of h2, yshift=-0.25cm] (a2) {$\boldsymbol{a}_2$}; %
 \node[obs, minimum size=0.4cm, above=of ht, yshift=-0.25cm] (at) {$\boldsymbol{a}_{t-1}$}; %
 \node[latent, minimum size=0.4cm, above=of ht1, yshift=-0.25cm, fill=yellow!30] (at1) {$\boldsymbol{a}_{t}$}; %
 \node[latent, minimum size=0.4cm, above=of ht2, yshift=-0.25cm, fill=yellow!30] (at2) {$\boldsymbol{a}_{t+1}$};
 \node[latent, minimum size=0.4cm, above=of hT, yshift=-0.25cm, fill=yellow!30] (aT) {$\boldsymbol{a}_{T}$}; %
 
 \node[obs, minimum size=0.4cm, below=of z1, yshift=0.25cm] (x1) {$\boldsymbol{x}_1$};
 \node[obs, minimum size=0.4cm, below=of z2, yshift=0.25cm] (x2) {$\boldsymbol{x}_2$};
 \node[obs, minimum size=0.4cm, below=of zt, yshift=0.25cm] (xt) {$\boldsymbol{x}_{t-1}$};
 \node[obs, minimum size=0.4cm, below=of zt1, yshift=0.25cm] (xt1) {$\boldsymbol{x}_{t}$};
 \node[latent, minimum size=0.4cm, below=of zt2, yshift=0.25cm] (xt2) {$\boldsymbol{x}_{t+1}$};
 \node[latent, minimum size=0.4cm, below=of zT, yshift=0.25cm] (xT) {$\boldsymbol{x}_{T}$};

  \node[obs, minimum size=0.4cm, below=of z1, yshift=0.35cm, xshift=-0.6cm, fill=green!30] (g1) {$\boldsymbol{g}$};
  \node[obs, minimum size=0.4cm, below=of z2, yshift=0.35cm, xshift=-0.6cm, fill=green!30] (g2) {$\boldsymbol{g}$};
  \node[obs, minimum size=0.4cm, below=of zt, yshift=0.35cm, xshift=-0.6cm, fill=green!30] (gt) {$\boldsymbol{g}$};
  \node[obs, minimum size=0.4cm, below=of zt1, yshift=0.35cm, xshift=-0.6cm, fill=green!30] (gt1) {$\boldsymbol{g}$};
  \node[obs, minimum size=0.4cm, below=of zt2, yshift=0.35cm, xshift=-0.6cm, fill=green!30] (gt2) {$\boldsymbol{g}$};
   \node[obs, minimum size=0.4cm, below=of zT, yshift=0.35cm, xshift=-0.6cm, fill=green!30] (gT) {$\boldsymbol{g}$};
 
   \edge {a1, h1, z1} {h2}
   \edge {at, ht, zt} {ht1}
   \edge {at1, ht1, zt1} {ht2}
   
   \edge {z1}{x1}
   \edge {z2}{x2}
   \edge {zt}{xt}
   \edge {zt1}{xt1}
   \edge {zt2}{xt2}
   \edge {zT}{xT}

   \edge {h1}{z1}
   \edge {h2}{z2}
   \edge {ht}{zt}
   \edge {ht1}{zt1}
   \edge {ht2}{zt2}
   \edge {hT}{zT}

   \edge {g1}{z1}
   \edge {g2}{z2}
   \edge {gt}{zt}
   \edge {gt1}{zt1}
   \edge {gt2}{zt2}
   \edge {gT}{zT}

   \path (h1) edge [bend left , ->] (x1);
   \path (h2) edge [bend left , ->] (x2);
   \path (ht) edge [bend left, ->] (xt);
   \path (ht1) edge [bend left , ->] (xt1);
   \path (ht2) edge [bend left , ->] (xt2);
   \path (hT) edge [bend left , ->] (xT);

   \path (x1) edge [densely dashed, bend left , ->] (z1);
   \path (h1) edge [densely dashed, bend right , ->] (z1);
   \path (x2) edge [densely dashed, bend left , ->] (z2);
   \path (h2) edge [densely dashed, bend right , ->] (z2);
   \path (xt) edge [densely dashed, bend left , ->] (zt);
   \path (ht) edge [densely dashed, bend right , ->] (zt);
   \path (xt1) edge [densely dashed, bend left , ->] (zt1);
   \path (ht1) edge [densely dashed, bend right , ->] (zt1);

   \path (g1) edge [densely dashed, bend left , ->] (z1);
   \path (g2) edge [densely dashed, bend left , ->] (z2);
   \path (gt) edge [densely dashed, bend left , ->] (zt);
   \path (gt1) edge [densely dashed, bend left , ->] (zt1);

   \draw [dotted] (2, 0) -- (2.3, 0);
   \draw [dotted] (1.7, -1.2) -- (2, -1.2);
   \draw [dotted] (2, 1.2) -- (2.3, 1.2);
   \draw [dotted] (2, 2) -- (2.3, 2);

   \draw [dotted] (6.7, 0) -- (7, 0);
   \draw [dotted] (6.4, -1.2) -- (6.6, -1.2);
   \draw [dotted] (6.7, 1.2) -- (7, 1.2);
   \draw [dotted] (6.7, 2) -- (7, 2);

 }
 
 \caption{Probabilistic graphical model of goal-conditioned recurrent state-space model. $\bm{x}_t$ and $\bm{z}_t$ respectively denote the observation and the latent state of the true state before step t, whilst $\bm{h}_t$ represents the hidden states of the recurrent backbone. $\bm{a}_t$ represents the action taken in step t, and $\bm{g}$ stands for the goal observation. The nodes coloured gray are the known inputs to the model, whilst the white ones are inferred or generated. The action nodes given in yellow indicates the ones sampled in the planning process.}
 \label{fig:rssm}
\end{figure}

Our \textit{LaGarNet} method extends \textit{PlaNet-ClothPick} by training a goal-conditioned latent dynamics model. Specifically, we extend the RSSM with goal-conditioned inference and generation, as shown in Figure~\ref{fig:rssm}, where the same goal observation $\bm{g}$ is incorporated into the estimation of the stochastic latent representation $\bm{z}$. The models are defined as:

\begin{align}
    \hat{\bm{z}}_t &\sim q(\hat{\bm{z}} \;|\; \bm{h}_t, \bm{x}_t, \bm{g}) 
    && \text{GC Representation Model} \label{eq:gc-repr} \\
    \tilde{\bm{z}}_t &\sim p(\tilde{\bm{z}} \;|\; \bm{h}_t, \bm{g}) 
    && \text{GC Transition Predictor} \label{eq:gc-trans}
\end{align}

Training objective of goal-conditioned RSSM (GC-RSSM) is to maximise log-likelihood of observations and rewards given sequence of actions and the target goal observation, i.e.,
\begin{equation}
    \argmax_{p_{\bm{x}}, p_{r}} \ln p_{\bm{x}}(\bm{x}_{1:T} | \bm{a}_{1:T}, \bm{g}) + \ln p_{r}(r_{1:T} | \bm{a}_{1:T}, \bm{g}) \label{eqn:obj}
\end{equation}

In addition, from the \text{probabilistic graphical model} of \text{RSSM} (Figure \ref{fig:rssm}), we know that
\begin{align}
    p(\bm{x}_{1:T}, \bm{z}_{1:T} | \bm{a}_{1:T}, \bm{g}) &= \prod_{t=1}^T p(\bm{x}_t|\bm{z}_t) p(\bm{z}_t|\bm{z}_{t-1}, \bm{a}_{t-1}, \bm{g}) \label{eqn:rep-1}\\
    q(\bm{z}_{1:T}| \bm{x}_{1:T}, \bm{a}_{1:T}, \bm{g}) &= \prod_{t=1}^T q(\bm{z}_t| \bm{x}_{1:t}, \bm{a}_{1:t}, \bm{g}) \label{eqn:rep-2},
\end{align}
where we ignore the deterministic latent states $\bm{h}_t$ as the direct product of it is the stochastic latent state $\bm{z}_t$. Expanding the objective \ref{eqn:obj} with the Equation \ref{eqn:rep-1} and \ref{eqn:rep-2}, we can obtain the loss function of GC-RSSM as
\begin{align}
    &\mathcal{L} = \sum_{t=1}^T \Bigg( - 
     \underset{\bm{z}_t}{\E} \Big[ \ln p(\bm{x}_t|\bm{z}_t) + \ln p(r_t|\bm{z}_t) \Big] \nonumber\\
    &+   \underset{\bm{z}_{t-1}}{\E}\Big[ KL\big[ q(\bm{z}_t|\bm{x}_{1:t}, \bm{a}_{1:{t-1}}, \bm{g}) || p(\bm{z}_{t}|\bm{z}_{t-1}, \bm{a}_{t-1}, \bm{g}   \big]\Big]
     \Bigg) \label{eqn:planet-objective-1}.
\end{align}

In  practice, the goal image $\bm{g} \in \R^{C\times H\times W}$ is conditioned on the stochastic latent states $\bm{z}$ in the form of goal-embedding $\bm{e}_g$ from the same encoder of current observation for our vision-based application. In training, we do not stop any gradients on the goals, but we stop them for overshooting loss. We use GC-RSSM to avoid recurrent structure of the model from ``forgetting'' information of the goal. 

We enlarge the network architecture to increase its capacity to address under-fitting issues on large diverse data. In contrast to \textit{PlaNet-ClothPick}, we increase the deterministic,  stochastic latent and hidden dimensions from (200, 30, 200) to (300, 60, 300). We also add two more linear layers in the reward prediction model.

\paragraph{Data Augmentation}
\text{Data Augmentation} techniques, such as scaling and rotation, on the observation are used to improve data efficiency for \text{end-to-end} cloth-shaping control systems \cite{lee2021learning}.  Domain randomisation, such as randomising the background colour of the table \cite{hoque2020visuospatial, yan2021learning, kadi2024mjtn}, is used for transferring the simulation-trained end-to-end policies to the real world. In this work, we employ vision processing for training to augment both the current and goal images to align the disparity between the synthetic and real observations \cite{kadi2024mjtn}. However, unlike the current image augmentation that uses trajectory-wise rotations and flipping, the augmentation on the goal image for input employs step-wise augmentation in addition to a translation using a Gaussian distribution $\mathcal{N}(0, 0.2^2)$. This augmented goal image is used for both input RGB (and/or depth) observation and output mask observation of the \text{RSSM} model. With such goal augmentation, we eliminate the need to provide a canonical goal state as input, which is particularly challenging to provide in real-world experiments. 

\subsection{Coverage-Alignment Reward}  \label{sec:ca-reward}

Our coverage-alignment reward $\hat{\mathcal{R}}_{CA}$  uses the step-wise coverage difference $\mathcal{R}_{dC}$, but it also employs the difference of maximum intersection-over-union (\text{Max IoU}) between the current and goal cloth mask, denoted as $\mathcal{R}_{du}$, to replace the the smoothness estimation in $\mathcal{R}_{CS}$, so that our method does not need to learn a smoothness prediction network as in \textit{SpeedFolding} \cite{avigal2022speedfolding}. Mathematically, our reward function that approximates the  \textit{SpeedFolding} reward can be expressed as follows: 

\begin{equation}
    \mathcal{R}_{SFA} = \max(\tanh(\alpha \times\mathcal{R}_{dc} + \beta\times \mathcal{R}_{du}), 0).
\end{equation}
In our experiments, we set $\alpha$ as 1 and $\beta$ as 2.  In order to encourage the maintenance of flattened states and discourage actions that disrupt nearly flattened states, similar to PlaNet-ClothPick's reward $\mathcal{R}_{PC}$, we propose the following augmented reward function based on the coverage-alignment reward:
\begin{equation}
    \hat{\mathcal{R}}_{CA} =
    \begin{cases}
      0 & NC - R_{dC} > 0.9 \textbf{ and } NC < 0.9  \\
      b  & NC \geq 0.95 \\
      \mathcal{R}_{SFA} & \textbf{otherwise}
    \end{cases},
\end{equation}
where we set $b = 0.7$ since the largest reward values of our reward lay around this value.

\subsection{Data Collection with Diffusion and Random Policies} \label{sec:data-collection}

\begin{figure}[t!]
    \centering
    \includegraphics[width=0.6\textwidth]{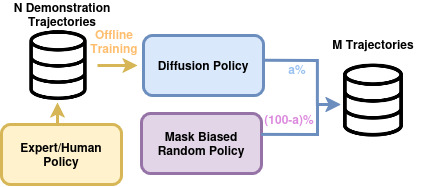}
    \caption{Data Collection of \textit{LaGarNet}. We first collection M human/expert demonstrations to train a \textit{Diffusion Policy}, then we use the mixture of the trained policy and the mask-biased random policy to collect trajectory data for LaGarNet.}
    \label{fig:data-collection}
\end{figure}

The major inductive biases introduced in the original \textit{PlaNet-ClothPick} come from its complex and specific blend of data collection policies. We also find that engineering flattening oracle policies for complex garment types, such as T-shirts and trousers, is an extremely challenging task. It defeats the simplicity advantage of imitation learning and also made offline reinforcement learning difficult to train as we cannot collect trajectories around the successful region of the distribution. Our new \textit{LaGarNet} method mitigates this bias using a \textit{Diffusion Policy} \cite{kadi2025draper, chi2023diffusion} and a mask-biased random policy (see Figure \ref{fig:data-collection}, Algorithm \ref{alg:data-collection}).

\begin{algorithm}
\caption{Data Collection with Diffusion and Random Policies for Offline Reinforcement Learning}
\label{alg:data-collection}
\KwIn{(1) Human/Expert demonstrations $\mathcal{D}_{\text{demo}}$,\\ (2) a random policy $\pi_{rdm}$,\\ (3) number of trajectories $M$ for collection, and \\ (4) ratio $a\%$ of diffusion trajectories}
\KwOut{Collected dataset $\mathcal{D}$ of trajectories}

\BlankLine
\Indp
Train a Diffusion Policy $\pi_{\text{diff}}$ on $\mathcal{D}_{\text{demo}}$ 
\Indm

\BlankLine
\For{$i = 1$ \KwTo $M$}{
    Sample $u \sim \text{Uniform}(0,1)$ \;
    \eIf{$u < a\%$}{
        Unroll trajectory $\tau$ using $\pi_{\text{diff}}$ \;
    }{
        Unroll trajectory $\tau$ using $\pi_{rdm}$ \;
    }
    $\mathcal{D}$.add($\tau$) \;
}
\BlankLine
\Return{$\mathcal{D}$}
\end{algorithm}

We train a \textit{Diffusion Policy} \cite{chi2023diffusionpolicy} using few human demonstrations to replace the oracle expert policies employed in \textit{PlaNet-ClothPick}. As shown in Table \ref{tab:longsleeve-flattening}, the \textit{Diffusion Policy} surprisingly generates numerous successful flattening trajectories; its action diversity accounts for some noisiness in manipulation, which can be utilised to diversify the data distribution around near-successful regions.

In addition, we also observe that it is unnecessary to use a completely random policy in our data collection process, as we employ a cloth mask to constrain the pick positions during planning using \textit{\text{model predictive control}}. Therefore, we propose a new mask-biased random policy, which uniformly selects a pick pixel on the cloth mask and a place pixel on the whole image, to replace completely random policies and the corner-biased random policy used in \textit{PlaNet-ClothPick}. By employing the \textit{Diffusion Policy} and mask-biased random policy, we significantly reduce the inductive biases present in the original algorithm.

\section{Action Inference  Transfer Heuristic for Robot Arm's Ring-Shaped Workspace} \label{sec:transfer-heuristic}


Many robot manipulators face workspace constraints, such as limited reach as well as undesirable trajectories caused by inverse kinematics and joint discontinuities. These factors often make parts of the workspace unusable for smooth pick-and-place (PnP) execution. To address this general class of problems, we propose a heuristic workspace sampling method that adapts trained policies to feasible regions of a robot’s workspace without retraining. We demonstrate this method on a \textit{UR3e} robot arm, whose PnP workspace resembles a ring due to both reach limitations and unstable trajectories near the base; the ring-like workspace is defined by a far radius $r_f$ and a near radius $r_n$ on the table surface.



\begin{figure}[t!]
    \centering
    \includegraphics[width=0.6\textwidth]{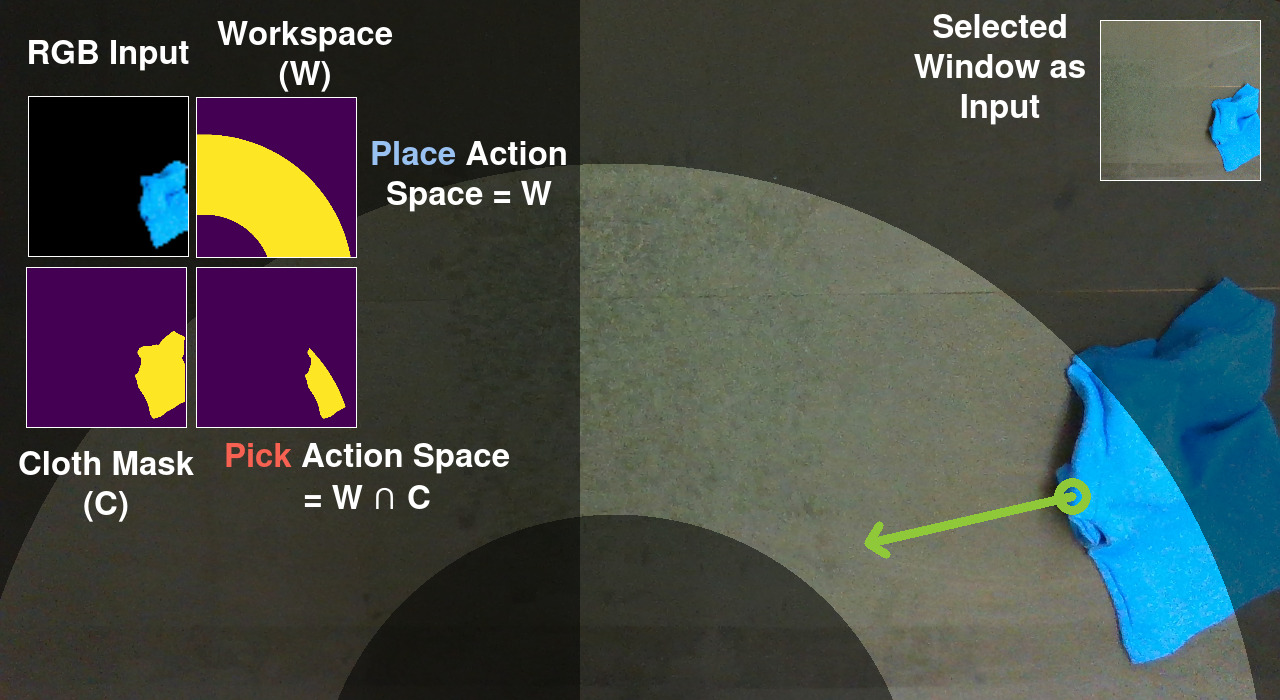}
    \caption[Heuristic of transferring policy trained in fixed square-widow as vision input to real-world ring-shape workspace of \textit{UR3e} robot arm.]{Heuristic of transferring policy trained in fixed square-widow as vision input to real-world constrained environment of \textit{UR3e} robot arm. The heuristic prioritises the selection of the middle window for network inputs, but if the selected window does not contain any parts of the cloth, it re-select a window that centres around the cloth as shown. The most highlighted part on the rectangular images are the workspace of the \textit{UR3e} Robot for a particular action step.}
     \label{fig:transfer-hueristic}
\end{figure}

Our heuristic workspace sampling method (see Figure \ref{fig:transfer-hueristic}) builds an observation bridge between the \textit{UR3e} environment and the policy's training environment. This method prioritises the central square window of the raw RGB input. If the central window does not contain any part of the cloth, it re-selects a window centred on the target cloth. The \textit{LaGarNet} latent dynamics network receives observations filtered by the cloth mask in the selected window, and the model predictive control planner samples pick actions at the intersection of the cloth mask and workspace mask, and place actions within the workspace mask in the selected window. Using these two processes, we ensure that \textit{LaGarNet} can be transferred to the entire visible workspace of a \textit{UR3e} robot from a top-down camera for flattening real garments. 

\section{Experiments} \label{sec:experiments}

We develop \textit{LaGarNet} using the garment simulation environment from Canberk et al.~\cite{canberk2023cloth}. Following the real-world adaptation strategy of the \textit{DRAPER} framework \cite{kadi2025draper}, the simulation includes extensions for real-world adaptation, specifically encompassing misgrasping and multi-layer grasping scenarios which are inherent to naive parallel grippers. 

We focus on simulated long-sleeved  T-shirts, trousers (pants), dresses and skirt objects selected by \cite{canberk2023cloth} from commonly employed \textit{Cloth3D} dataset \cite{bertiche2020cloth3d} (see Figure \ref{fig:clothfunnels-garments}). Following Canberk et al.~\cite{canberk2023cloth}, we also scale the garments by a factor of 0.8 and set the simulated camera height as 2m above the surface with a completely dark  background for the vision rendering. We set the stretch, bend and shear stiffness of the cloth meshes as (0.75, .02, .02) and cloth mass as 0.5kg for all garments. Canberk et al.~\cite{canberk2023cloth} split the generated initial garments into  ``hard'' and ``easy'' categories where the former is more crumpled and the latter is more flattened. We use 30 ``hard'' initial states in the evaluation objects placed within the defined workspace of the gripper for evaluation.




\subsection{Comparison of \textit{LaGarNet} against Other Neural Controllers}

\begin{table*}[t!]
\caption{Numerical comparison of different learning-based pick-and-place (\text{PnP}) controllers on longsleeved T-shirts in simulation. We adopt normalised coverage (\text{NC}, \%), normalised improvement (\text{NI}, \%), and maximum intersection-over-union (\text{Max IoU}, \%) between the current and goal cloth mask to evaluate the performance of flattening. SR denotes the flattening success rate; a successful trajectory reaches 90\% \text{NC} and 80\% \text{Max IoU} within a defined steps. We only enable the PnP primitives of the off-the-shelf \textit{ClothFunnels} \cite{canberk2023cloth}. Both \textit{PlaNet-ClothPick} and \textit{LaGarNet} are trained with the RGB-to-Mask setup on the same task-focused and all-garments datasets, yet with their own reward functions and network architectures. The all-garments dataset is four times larger than the task-focused dataset, as it includes data from four garment types, while the task-focused dataset contains only longsleeve data. We train the behaviour cloning methods \textit{JA-TN} \cite{kadi2024mjtn} and \textit{Diffusion Policy} \cite{kadi2025draper, chi2023diffusionpolicy} with 50 human demonstrations. Mesh-based planning method, $\dagger$ VCD \cite{lin2020softgym} , reports to reach  83.8 $\pm$ 45 \% NI and 96 $\pm$ 21.8 \% NC for T-shirt flattening in simulation within 20 steps; its improved successor, $\ddagger$ MEDOR, reports to achieve above 90\% NI within 10 steps. The proposed all-garments \textit{LaGarNet} outperforms all compared algorithms, but it does not reach human-level performance.}
\label{tab:longsleeve-flattening}
\centering
\resizebox{\textwidth}{!}{%
\begin{tabular}{l|cccp{2.1cm}p{2.1cm}p{2.1cm}p{2.1cm}||c}
Longsleeve & ClothFunnels & JA-TN & Diffusion Policy & PlaNet-ClothPick
(Task Focused) & PlaNet-ClothPick
(All Garment) & LaGarNet
(Task Focused) & LaGarNet
(All Garments) & Human\\
\hline
NI $\uparrow$ (10 steps) &42.7 $\pm$ 19.5 & 30.7 $\pm$ 26.8 & 54.5 $\pm$ 21.2 & 47.2 $\pm$ 22.7 & 48.6 $\pm$ 19.3 & 66.2 $\pm$ 23.4 & \underline{73.6 $\pm$ 15.9} $\ddagger$ & 79.8 $\pm$ 19.5 \\
NC $\uparrow$&74.0 $\pm$ 11.3 & 65.7 $\pm$ 15.9 & 79.1 $\pm$ 8.8 & 76.0 $\pm$ 9.3 & 75.9 $\pm$ 10.6 & 83.3 $\pm$ 12.3 & 86.8 $\pm$ 10.3 & 90.5 $\pm$ 10.3 \\
Max IoU $\uparrow$&53.2 $\pm$ 13.9 & 53.5 $\pm$ 11.1 & 61.9 $\pm$ 9.0 & 59.9 $\pm$ 10.1 & 60.3 $\pm$ 7.5 & 66.8 $\pm$ 12.6 & 69.4 $\pm$ 9.8 & 74.3 $\pm$ 10.6 \\
SR $\uparrow$&0/30 & 1/30 & 2/30 & 3/30 & 0/30 & 7/30 & 6/30 & 14/30 \\
\hline
NI $\uparrow$ (20 steps)&48.6 $\pm$ 21.3 & 50.2 $\pm$ 28.0 & 68.2 $\pm$ 18.0 & 49.0 $\pm$ 20.4 & 52.7 $\pm$ 17.5 & 75.5 $\pm$ 21.1 & \underline{79.5 $\pm$ 16.5} $\dagger$ & 89.1 $\pm$ 11.0 \\
NC $\uparrow$&76.9 $\pm$ 11.3 & 75.6 $\pm$ 15.7 & 85.4 $\pm$ 7.9 & 76.7 $\pm$ 8.7 & 78.0 $\pm$ 9.2 & 88.3 $\pm$ 10.2 & \underline{89.8 $\pm$ 9.7} $\dagger$  & 95.4 $\pm$ 4.6 \\
Max IoU $\uparrow$&54.6 $\pm$ 14.3 & 60.2 $\pm$ 11.5 & 67.5 $\pm$ 10.1 & 61.8 $\pm$ 8.9 & 61.4 $\pm$ 7.8 & 72.5 $\pm$ 11.5 & 74.7 $\pm$ 11.2 & 80.7 $\pm$ 4.0 \\
SR $\uparrow$&0/30 & 3/30 & 5/30 & 3/30 & 1/30 & 14/30 & 18/30 & 22/30 \\
\hline
NI  $\uparrow$ (30 steps) &54.6 $\pm$ 22.0 & 42.3 $\pm$ 32.5 & 78.7 $\pm$ 17.8 & 49.8 $\pm$ 22.2 & 53.4 $\pm$ 18.0 & 78.8 $\pm$ 17.2 & \textbf{87.2 $\pm$ 11.6} & 91.3 $\pm$ 8.7 \\
NC $\uparrow$&79.1 $\pm$ 11.9 & 70.5 $\pm$ 18.4 & 89.9 $\pm$ 8.7 & 77.2 $\pm$ 9.7 & 78.5 $\pm$ 8.8 & 89.9 $\pm$ 8.8 & \textbf{94.2 $\pm$ 6.1} & 96.5 $\pm$ 3.3 \\
Max IoU $\uparrow$&55.9 $\pm$ 15.2 & 57.1 $\pm$ 14.2 & 70.7 $\pm$ 10.9 & 62.4 $\pm$ 9.1 & 61.9 $\pm$ 8.1 & 72.5 $\pm$ 12.9 & \textbf{79.5 $\pm$ 7.3} & 81.9 $\pm$ 3.2 \\
SR $\uparrow$&2/30 & 4/30 & 11/30 & 3/30 & 2/30 & 17/30 & \textbf{23/30} & 29/30 \\
\bottomrule
\end{tabular}
}

\end{table*}

We benchmark our model-based reinforcement learning (RL) method \textit{LaGarNet} against its previous counterpart \textit{PlaNet-ClothPick} \cite{kadi2024planet} and the off-the-shelf model-free method \textit{ClothFunnels} \cite{canberk2023cloth} for garment-flattening with its pick-and-place (PnP) primitives.  We also compare LaGarNet against the advanced deep behaviour cloning method \textit{JA-TN} \cite{kadi2024mjtn} and a modified \textit{Diffusion Policy} for PnP cloth-shaping domain. All trained methods apart from \textit{ClothFunnels} employ the vision processing strategy used by \textit{DRAPER} framework to achieve simulation-to-reality transfer \cite{kadi2024mjtn}; this strategy attempts to align the real and synthetic RGB-D observations, apply real-world adaption and domains randomisation in simulation as well as a refined grasping procedure to reduce grasping failures. Whilst we train  \textit{JA-TN} and \textit{Diffusion Policy} on 50 human expert trajectories, we let task-focused \textit{LaGarNet} and \textit{PlaNet-ClothPick} learn from 300 thousand transitional steps generated from proposed data-collection strategy. We also develop all-garment variants of these methods using trajectories from all four garment types. Human experiments are conducted through a PnP window interface operating directly on observation images both in simulation and the real world. 

\begin{table*}[t!]
\caption{Real-world garment flattening performance of different pick-and-place (PnP) controllers.  We adopt normalised coverage (\text{NC}, \%), normalised improvement (\text{NI}, \%), and maximum intersection-over-union (\text{Max IoU}, \%) between the current and goal cloth mask to evaluate the performance of flattening. SR$^{90}_{80}$  denotes the flattening success rate; a successful trajectory reaches 90\% \text{NC} and 80\% \text{Max IoU} within a defined steps; we also use SR$^{90}_{0}$ represents a trajectory success where only NC reaches 90\%. Both \textit{PlaNet-ClothPick} and \textit{LaGarNet} were trained on identical garments using the same RGB-to-Mask variant and deployed via the \textit{DRAPER} \cite{kadi2025draper} real-world cloth manipulation framework. Mesh-based planning method, $\dagger$ VCD \cite{lin2020softgym} , reports to reach 79.2 $\pm$ 9.4 \% NI and  77.3  $\pm$ 14.1 \% NC for real T-shirt flattening within 20 steps; its improved successor, $\ddagger$ MEDOR, reports to achieve around 64\% NI flattening for real trousers and 46.8\% NI for real dresses within 10 steps. \textit{LaGarNet} shows substantial performance improvements across all garments compared to \textit{PlaNet-ClothPick}, though it still falls short of human-level performance in PnP garment flattening.}
\label{tab:real-worldflattening}
\centering
\resizebox{\textwidth}{!}{%
\renewcommand{\arraystretch}{1.2} 
\setlength{\tabcolsep}{2pt} 
\begin{tabular}{c|ccccc|ccccc||ccccc}
Method & \multicolumn{5}{c|}{PlaNet-ClothPick \cite{kadi2024planet}}  & \multicolumn{5}{c||}{LaGarNet} & \multicolumn{5}{c}{Human Policy} \\
\cline{1-16}
10 Steps & NC $\uparrow$  & NI $\uparrow$ & Max IoU $\uparrow$    & SR$^{90}_{80}$  $\uparrow$  & SR$^{90}_{0}$  $\uparrow$ 
& NC $\uparrow$  & NI $\uparrow$ & Max IoU $\uparrow$     & SR$^{90}_{80}$  $\uparrow$  & SR$^{90}_{0}$  $\uparrow$ 
& NC $\uparrow$  & NI $\uparrow$ & Max IoU $\uparrow$    & SR$^{90}_{80}$  $\uparrow$  & SR$^{90}_{0}$  $\uparrow$ \\
\hline
Longsleeve & 62.1 $\pm$ 9.6 & 23.7 $\pm$ 22.5 & 58.2 $\pm$ 6.5 & 0/10 & 0/10 & 82.7 $\pm$ 16.7 & 73.0 $\pm$ 19.9 & 72.9 $\pm$ 13.5 & 5/10 & 5/10 & 91.2 $\pm$ 14.5 & 84.0 $\pm$ 26.3 & 83.1 $\pm$ 13.5 & 8/10 & 8/10 \\
Trousers & 67.9 $\pm$ 12.9 & 38.3 $\pm$ 24.2 & 60.8 $\pm$ 11.1 & 0/10 & 0/10 & 81.1 $\pm$ 9.2 & \underline{66.1 $\pm$ 14.6} $\ddagger$ & 74.9 $\pm$ 8.7 & 1/10 & 1/10 & 91.7 $\pm$ 10.0 & 85.1 $\pm$ 18.4 & 83.8 $\pm$ 9.2 & 7/10 & 8/10 \\
Dress & 58.2 $\pm$ 11.9 & 22.2 $\pm$ 15.4 & 54.4 $\pm$ 10.2 & 0/10 & 0/10 & 76.5 $\pm$ 6.4 & \underline{59.1 $\pm$ 12.8} $\ddagger$ & 67.6 $\pm$ 5.1 & 0/10 & 0/10 & 73.6 $\pm$ 14.6 & 48.8 $\pm$ 27.5 & 66.8 $\pm$ 12.8 & 2/10 & 2/10 \\
Skirt & 70.4 $\pm$ 16.3 & 42.3 $\pm$ 33.3 & 68.6 $\pm$ 14.2 & 2/10 & 2/10 & 92.8 $\pm$ 3.1 & 83.1 $\pm$ 8.5 & 87.8 $\pm$ 2.6 & 10/10 & 10/10 & 96.2 $\pm$ 4.5 & 92.2 $\pm$ 9.7 & 90.2 $\pm$ 3.8 & 9/10 & 9/10 \\
\hline
Real Total & 64.6 $\pm$ 13.3 & 31.6 $\pm$ 25.4 & 60.5 $\pm$ 11.7 & 2/40 & 2/40 & 83.3 $\pm$ 11.5 & 70.3 $\pm$ 16.6 & 75.8 $\pm$ 11.1 & 16/40 & 16/40 & 88.2 $\pm$ 14.2 & 77.5 $\pm$ 26.9 & 81.0 $\pm$ 13.4 & 26/40 & 27/40 \\

\hline
\hline

20 Steps 
& NC $\uparrow$  & NI $\uparrow$ & Max IoU $\uparrow$    & SR$^{90}_{80}$  $\uparrow$  & SR$^{90}_{0}$  $\uparrow$ 
& NC $\uparrow$  & NI $\uparrow$ & Max IoU $\uparrow$     & SR$^{90}_{80}$  $\uparrow$  & SR$^{90}_{0}$  $\uparrow$ 
& NC $\uparrow$  & NI $\uparrow$ & Max IoU $\uparrow$    & SR$^{90}_{80}$  $\uparrow$  & SR$^{90}_{0}$  $\uparrow$ \\
\hline
Longsleeve & 64.9 $\pm$ 9.9 & 29.0 $\pm$ 24.3 & 60.1 $\pm$ 7.1 & 0/10 & 0/10 & \underline{87.7 $\pm$ 9.8} $\dagger$ & \underline{79.2 $\pm$ 16.4} $\dagger$ & 77.6 $\pm$ 9.2 & 6/10 & 6/10 & 97.3 $\pm$ 3.1 & 95.0 $\pm$ 5.5 & 88.5 $\pm$ 5.6 & 9/10 & 10/10 \\
Trousers & 72.2 $\pm$ 13.0 & 46.8 $\pm$ 23.6 & 64.2 $\pm$ 11.2 & 0/10 & 0/10 & 85.4 $\pm$ 7.6 & 74.2 $\pm$ 11.8 & 77.9 $\pm$ 6.4 & 1/10 & 2/10 & 96.3 $\pm$ 6.8 & 93.4 $\pm$ 12.2 & 88.3 $\pm$ 7.3 & 9/10 & 9/10 \\
Dress & 60.6 $\pm$ 12.9 & 26.8 $\pm$ 17.9 & 55.5 $\pm$ 10.5 & 0/10 & 0/10 & 84.9 $\pm$ 6.5 & 73.9 $\pm$ 11.6 & 73.3 $\pm$ 4.0 & 0/10 & 2/10 & 86.8 $\pm$ 8.7 & 73.8 $\pm$ 16.8 & 76.9 $\pm$ 9.8 & 4/10 & 5/10 \\
Skirt & 77.1 $\pm$ 14.2 & 55.1 $\pm$ 29.1 & 74.6 $\pm$ 11.8 & 3/10 & 3/10 & 92.8 $\pm$ 3.1 & 83.1 $\pm$ 8.5 & 87.8 $\pm$ 2.6 & 10/10 & 10/10 & 96.3 $\pm$ 4.4 & 92.5 $\pm$ 9.6 & 90.4 $\pm$ 3.6 & 9/10 & 9/10 \\
\hline
Real Total & 68.7 $\pm$ 13.7 & 39.4 $\pm$ 26.1 & 63.6 $\pm$ 12.2 & 3/40 & 3/40 & 87.7 $\pm$ 7.6 & 77.6 $\pm$ 12.5 & 79.2 $\pm$ 8.0 & 17/40 & 20/40 & 94.2 $\pm$ 7.3 & 88.7 $\pm$ 14.3 & 86.0 $\pm$ 8.6 & 31/40 & 33/40 \\
\bottomrule
\end{tabular}

}

\end{table*}

\begin{figure}[t!]
    \centering
    
    \subfloat[\centering Step 20 \label{fig:goal-ablation}] {{\includegraphics[width=0.48\textwidth]{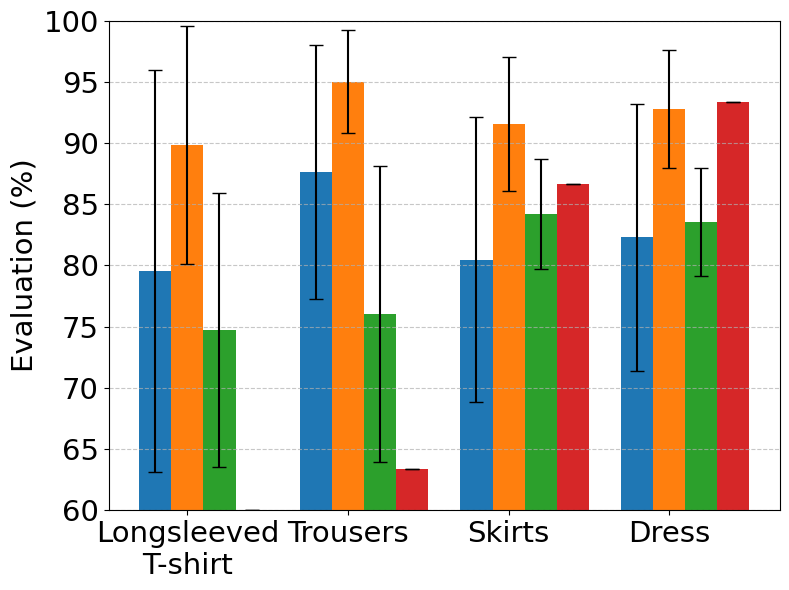} }}
    \subfloat[\centering Step 30  \label{fig:reward-abaltion}] {{\includegraphics[width=0.48\textwidth]{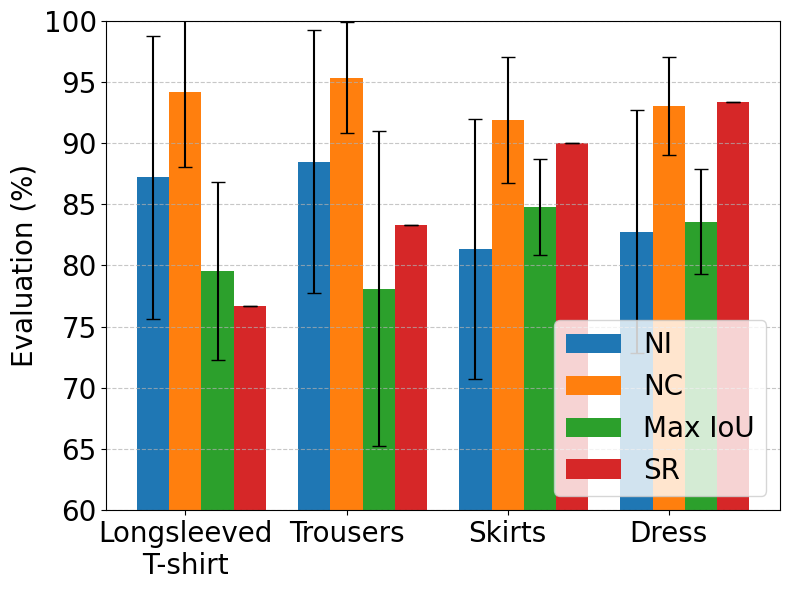} }}
    
    \caption{Simulation performance of single-policy LaGarNet on all four garment types. LaGarNet demonstrates higher performance on skirts and dresses compared to longsleeved T-shirts and trousers, but it achieves above 90\% NC flattening performance across all garments. However, it struggles to achieve higher Max IoU than 85\%.
    } \label{fig:sim-lagarnet-results}
\end{figure}

Table \ref{tab:longsleeve-flattening} indicates that both task-focused and all-garment versions of \textit{LaGarNet} perform much better than its original counterpart \textit{PlaNet-ClothPick} and \textit{ClothFunnels}'s \text{PnP} policy, but do not quite match the human policy. \textit{Diffusion Policy} performs  substantially better than another behaviour cloning method \textit{JA-TN} whilst training on the same amount of human trajectories and the same update steps, but it cannot reach the performance of \textit{LaGarNet}, possibly due to the amount of available data. In addition, all methods struggle to reach high \text{Max IoU} values. Figure \ref{fig:sim-lagarnet-results} shows that single-policy LaGarNet can achieve above 90\% NV on all examined garments in simulation, but the learning seems like overfitting to dresses and skirts as indicated by the success rate.

\subsection{Transfer Ability of LaGarNet to Real-World Settings}

Our real-world study assess the same all-garment \textit{LaGarNet} and \textit{PlaNet-ClothPick} policy examined in the last section on on all four types of garments; see Figure \ref{fig:real-garments} for the colour, size and material information of these garments.

Following Section \ref{sec:transfer-heuristic}, workspace of our \textit{UR3e} robot is defined by a far radius $r_{f}$ of 0.54cm and a near radius $r_{n}$ of 0.24cm, and the camera can only see the front-half segment of it in our setup as show in Figure \ref{fig:transfer-hueristic}. For a top-down camera, we can find the space position $p_w$ in the world frame of each pixels $p_x$ on the camera view using the camera intrinsic and the transformation matrix of the camera. We set the \textit{RealSense D435i} camera  (with field-of-view of 69$^\circ$ $\times$ 42$^\circ$ and resolution of 720$\times$1280) 0.72m above the table in our real-world setup. Finally, we can decide if $p_x$ lies in the workspace by checking if $r_n \leq ||p_w - b_w||_2 \leq r_f$, where $b_w$ represent the position of the robot base in the world frame. We replace the tweezer extension previously used in \textit{DRAPER} \cite{kadi2025draper} and \textit{LazyDagger} \cite{hoque2021lazydagger} with a 3D-printed cylinder extension to improve the holding capacity for heavier garments.  

Table \ref{tab:real-worldflattening} shows that the all-garment version of \textit{LaGarNet} can flatten longsleeved T-shirts, trousers, and skirts in the real world, but it struggles with dresses. Human policy also struggles to flatten the  dress because it is too soft, making it hard to separate the sleeves from the garment body. \textit{LaGarNet} performs substantially better than \textit{PlaNet-ClothPick}. Notably, the NC and NI performance of \textit{LaGarNet} reaches state-of-the-art levels for single-gripper garment flattening when compared to mesh-based planning methods \cite{lin2022learning, huang2022mesh}; the reported values are mentioned in Section \ref{sec:mbrl}, demonstrating near-human performance on garment flattening with a state-space model

\subsection{Ablation Study on \textit{LaGarNet}}

\begin{figure}[t!]
    \centering
    
    \subfloat[\centering Goal Ablation \label{fig:goal-ablation-1}] {{\includegraphics[width=0.5\textwidth]{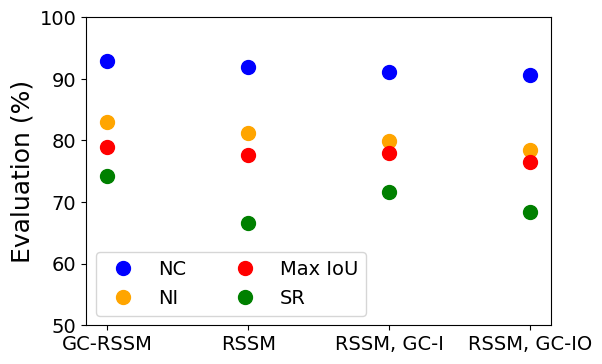} }}
    \subfloat[\centering Reward Ablation\label{fig:reward-abaltion-1}] {{\includegraphics[width=0.45\textwidth]{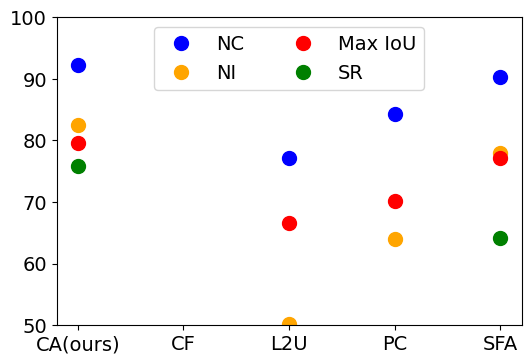} }}
    
    \caption{Ablation study on \textit{LaGarNet} in simulation environments. We use the all-garment RGB-to-RGB \textit{LaGarNet} model in this study. Results are aggregated across 4 garment types $\times$ 30 evaluation trials per type. For reward function ablation, we train \textit{LaGarNet} with five reward variants: Coverage-alignment (CA, ours), PlaNet-ClothPick (PC) \cite{kadi2024planet}, ClothFunnels (CF), Learning2Unfold (L2U) \cite{gu2023learning}, SpeedFolding approximation (SFA). Missing evaluation points indicate values below 50\%. Our proposed GC-RSSM and reward function is eseential to the success of \textit{LaGarNet}.
    } \label{fig:abalation-task-focused}
\end{figure}

\begin{table}[t!]
\centering
\caption{Quantitative results for RGB image reconstruction quality from various world models are presented. We evaluate the visual quality and similarity of generated images using Mean Squared Error (MSE) and Structural Similarity Index (SSIM) \cite{wang2004image}. SSIM is especially useful as it evaluates local structure and contrast, which are directly affected by blurring. All values are multiplied by 100 for clearer comparison. Each value represents the aggregate average over 10 trials for each of the 4 garments. All these models are based on same internal hyperparameters; the differences lie in how the models are structured and how their inputs and outputs differ. Our goal-conditioned recurrent state-space model (GC-RSSM) shows a improvement of reducing blurriness on the multi-step prior reconstructed images.
}
\label{tab:recon-quality}
\resizebox{0.6\textwidth}{!}{
\setlength{\tabcolsep}{2pt} 
\renewcommand{\arraystretch}{1.1} 

\begin{tabular}{l|cc|cc|cc|cc}
 & \multicolumn{2}{c|}{Posterior} & \multicolumn{2}{c|}{Prior 1} & \multicolumn{2}{c|}{Prior 2} & \multicolumn{2}{c}{Prior 3} \\
\hline
 Model, Metrics & MSE & SSIM   & MSE & SSIM   & MSE & SSIM   & MSE & SSIM  \\
\hline
RSSM & \textbf{0.26} & \textbf{94.09} &1.92 & 63.37 &2.60 & 62.00 &2.91 & 52.17\\
RSSM, GC-I & 0.28 & 94.72 &1.87 & 65.58 &2.59 & 58.46 &2.80 & 56.05\\
RSSM, GC-IO & 0.37 & 92.78 & \textbf{1.57} & \textbf{68.12} & \textbf{2.34}  & 62.10 &2.70 & 54.17\\
GC-RSSM & 0.32 & 94.03 &1.67 & 67.72 &2.40 & \textbf{63.27} & \textbf{2.70} & \textbf{57.66}\\
\bottomrule
\end{tabular}
}
\end{table}

\begin{figure}[t!]
    \centering
    \includegraphics[width=0.8\linewidth]{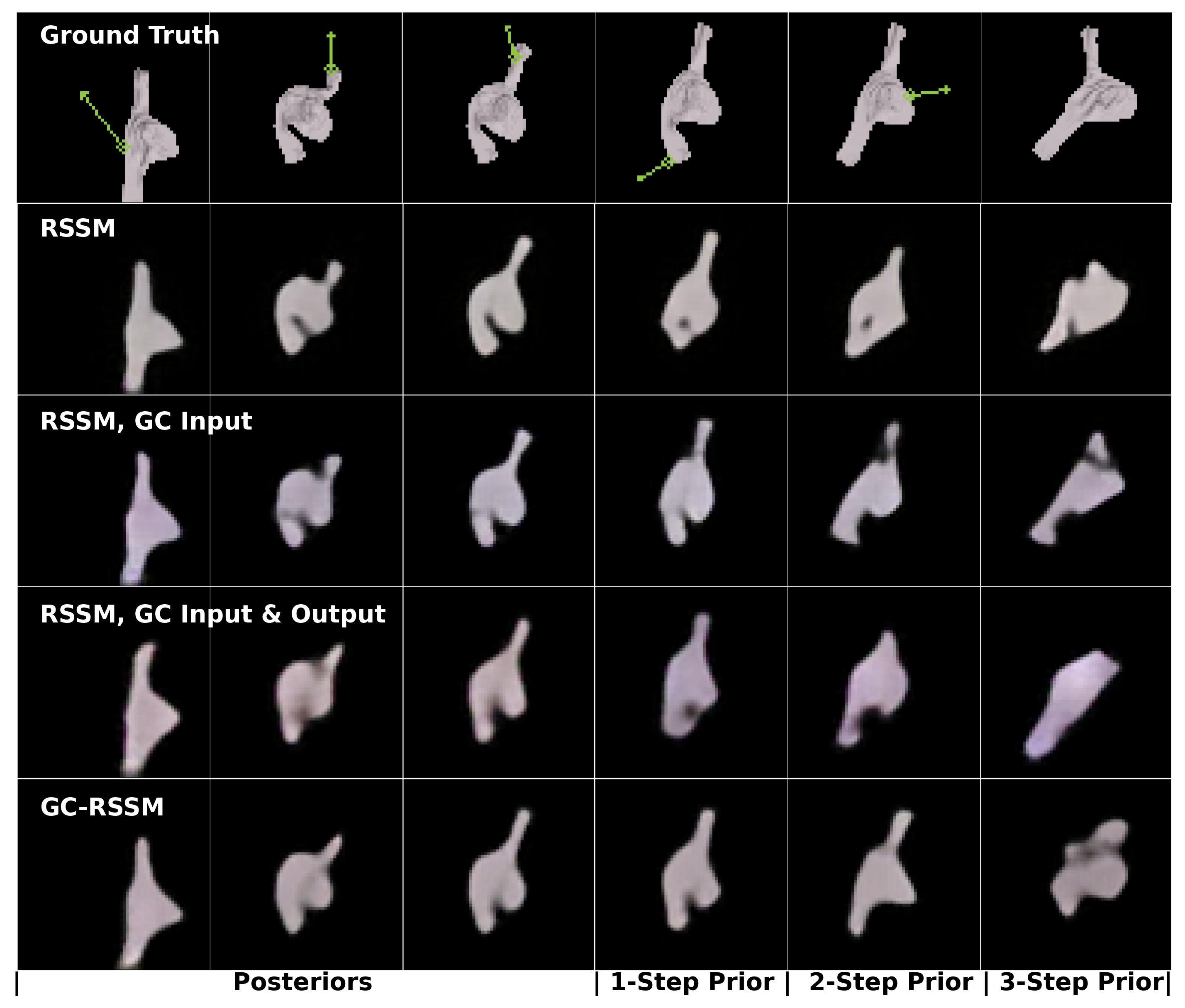}
    \caption{Observation reconstruction quality of various world models; each grid represents a 64$\times$64 RGB image. The green arrows in the first row represent the pick-and-place actions on each step. Our GC-RSSM model demonstrates a superior ability to track the sleeves of the garment in prior steps compared to other RSSM variants.}
    \label{fig:recon-quality}
\end{figure}

This section examines the effect of different components of \textit{LaGarNet}, including data collection, reward function and the goal-condition recurrent state-space model (GC-RSSM).

Figure \ref{fig:recon-quality} and Table \ref{tab:recon-quality} show that our GC-RSSM model provides a superior prior latent state representation, which is crucial for improved planning and, consequently, better control performance as demonstrated in Figure \ref{fig:goal-ablation}. In addition, Figure \ref{fig:reward-abaltion} shows that the proposed coverage-alignment (CA) reward function is essential to the success of the policy. Lastly, comparing the performance of our reward function and the SpeedFolding approximation reward (SFA) \cite{avigal2022speedfolding}, we find that augmented reward that encourages almost flattened steps and discourages messing up the flattened steps is really effective, but we have to set the bonus value which needs lie in the reward distribution.

\begin{figure}[t!]
    \centering
    \includegraphics[width=0.5\textwidth]{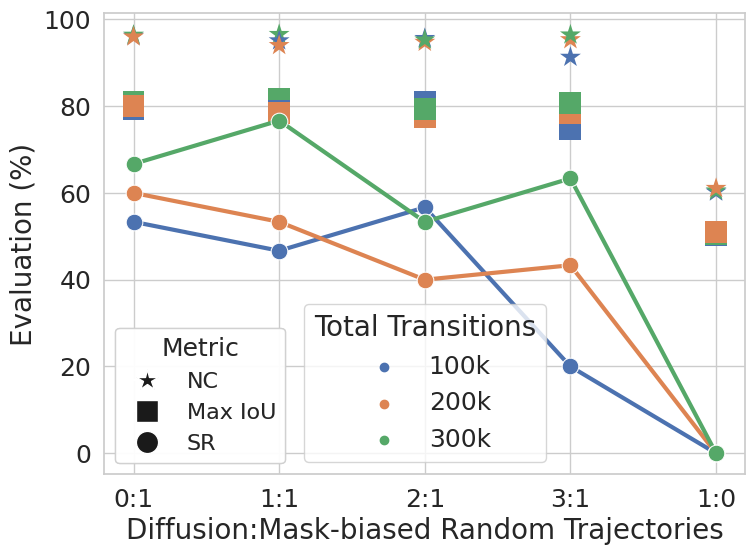}
    \caption{Ablation of Data Collectoin. We train the RGB-to-RGB variant of \textit{LaGarNet} on task-specific datasets of various sizes and different trajectory ratios between the \textit{Diffusion Policy} and mask-biased random Policy. SR represents the success rate, where a successful trajectory reaches 90\% normalised coverage (\text{NC}) and 80\% maximum intersection-over-union (\text{Max IoU}) within 30 steps.   In general, more data leads to better flattening performance. A mask-biased random policy is significant for the success of LaGarNet, while a diffusion policy can further boost performance when a larger amount of data is available.}
    \label{fig:data-collection-ablation}
\end{figure}

Figure \ref{fig:data-collection-ablation} indicates that a mask-biased random policy is essential for the success of \textit{LaGarNet}, and that \textit{Diffusion Policy} can further boost performance when a large amount of data is available. Although the ratio between the two policies does not affect the NC and Max IoU values, the success rate is quite sensitive to this ratio.

\section{Discussion} \label{sec:discussion}

The key reason for our investigation into latent dynamic representations is their potential to achieve generalisability across a broader range of contexts. By leveraging such representations, systems developed in this study may ultimately be extended beyond laboratory environments and applied in real-world scenarios, including everyday life and other discussed diverse settings. There are two critical challenges in applying RSSM in pick-and-place cloth shaping: (1) capturing the complex deformation of cloth objects and (2) quasi-static nature of high-level action primitive -- the inferred states are not as  continuous as the ones in velocity/position-based control. We find that \text{RSSM} \cite{hafner2019learning} with its goal-conditioned extension is better at learning complex latent dynamics of cloth deformation caused by higher-level action primitives.

In general, offline reinforcement learning suffers from the problem of distribution shift \cite{prudencio2023survey}, where the testing trajectories misalign with those collected during training. Most of the offline RL literature relies on datasets collected using a single policy \cite{agarwal2020optimistic}. We find that the behaviour cloning method, \textit{Diffusion Policy} \cite{chi2023diffusionpolicy}, can learn to flatten garments from only 50 human demonstrations and can be used to collect data in the near-success distribution for offline RL algorithms, as it produces diverse actions around the expert-trajectory distribution. Building on this, our paper proposes a potentially generalisable data collection strategy for offline RL, which combines both a random policy and a learned diffusion policy from a view demonstration approach.

\subsection{Goal-Conditioned Methods using RSSM}
Some previous research attempted to integrate goal-conditioned learning in RSSMs. Pertsch et al.~ \cite{pertsch2020long} proposed a goal-condition predictor \textit{(GCP)} that conditioned on the initial and goal states of subtask of a long horizon tasks, and they further improve the planning by using a hierarchical divide-and-conquer scheme; the action is inferred inversely from current and an optimised sub-goal. Duan et al. ~\cite{duan2024learning} present a goal-directed exploration algorithm, \textit{MUN}, and they also devise \textit{GC-Dreamer} as a baseline to compare against \textit{MUN}; but the goal of \textit{GC-Dreamer} is conditioned on the actor and the critic network instead of the world model. Our method is different from these method because we condition the goal directly on the latent state inference of RSSM for learning better representation.

\subsection{Limitations and Failure Cases}

\begin{figure}[t!]
    \centering
    \includegraphics[width=0.8\linewidth]{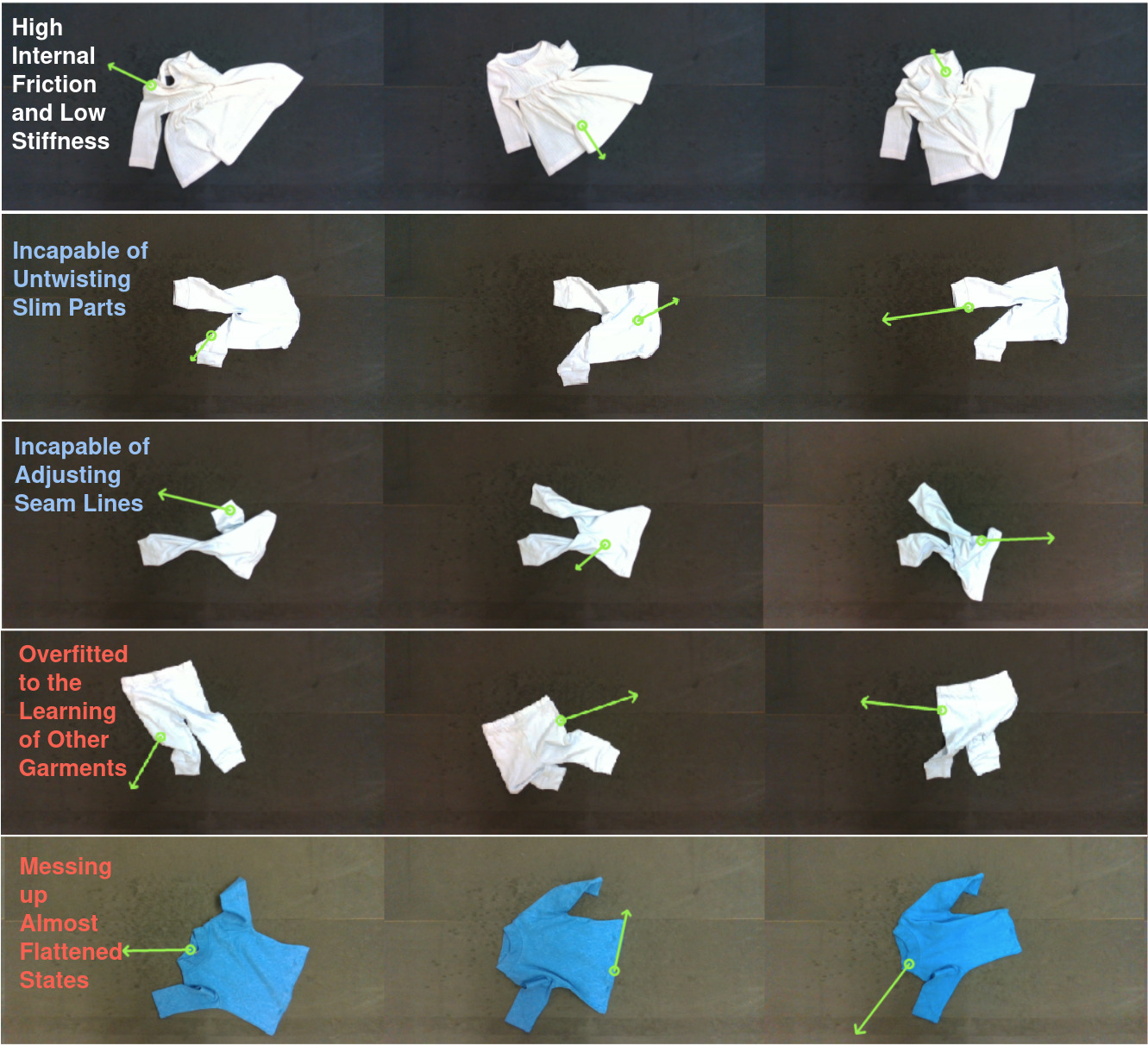}
    \caption{Failure cases of LaGarNet in real-world experiments. We highlight failures due to mechanical limitations with white text, those due to assumption oversights with blue text, and those due to the learning process itself with red text.}
    \label{fig:failure}
\end{figure}

Despite the substantial successes of \textit{LaGarNet} demonstrated in the last section, it encounters significant challenges in real-world applications. We conduct our study under the assumption that intrinsic physical properties of the target cloth, such as stiffness and elasticity, are less critical for high-level policy learning. As a result, the system struggles to flatten garments with high internal friction and small stiffness due to its single-arm as shown in the top row of Figure \ref{fig:failure}, parallel gripper design --- a challenge that even humans face when attempting to unfold stubborn folds with a single gripper as discussed in the last section. 

Additionally, \textit{LaGarNet} has difficulty in flattening trousers and skirts with wide openings, as it cannot reliably identify and adjust seam lines as shown in the third row of Figure \ref{fig:failure}. The policy also tends to disrupt nearly flattened garments, and it fails to accurately select and untwist slim parts of garments (see the second row of Figure \ref{fig:failure}), like sleeves and legs. Furthermore, since distinguishing the front from the back of garments was not included as a task criterion in our system, \textit{LaGarNet} is inherently unable to perform this distinction --- a task that is generally considered to be challenging even for young human children \cite{paoletti2012pink}.

Furthermore, we also encounter classical machine learning challenges. When training on all-garment datasets, we observe general underfitting issues compared to task-specific models; specifically, the all-garment policy tends to underfit the longsleeved T-shirts and trousers trajectories while overfitting to the other two garment types, as reflected in poorer performance on T-shirts and trousers (see the fourth row of Figure \ref{fig:failure}). To address this, it is important to balance the ratio of trajectories generated by the expert diffusion policy and the mask-biased random policy, ensuring the policy neither overfit nor underfit particular trajectory types. Increasing model capacity can help alleviate general underfitting as we did in this paper, while prioritised sampling can correct internal imbalances in fitting across garment types. In our experiment, we set the diffusion to random ratio as 1:1, but more generalisable approach should be developed in the future. Lastly, our GC-RSSM latent dynamic model still cannot accurate predict the first-step prior reward consistently, causing it generate action that messes up garment in almost flattened states as show in the last row of Figure \ref{fig:failure}.

\subsection{Future}

We intend to extend \textit{LaGarNet} to achieve garment folding tasks with an improved learning and inference capability. We also aim to explore other action primitives, such as bimanual \text{pick-and-fling}, \text{pick-and-drag}, and hybrids of these actions \cite{avigal2022speedfolding,canberk2023cloth} to improve the versatility and efficiency of our system.  We also curious about the effectiveness of the proposed coverage-alignment reward in learning from real-world data. We also plan to investigate a transformer-based architecture \cite{deng2023facing} to replace the recurrent structure of our world model.

The findings of the paper indicate that learning-based methods remain inferior to human-level manipulation under the same constraints. We speculate that one advantage humans possess is a strong understanding of the topological structures of various garment types. This naturally leads to new research questions: \textit{can neural controllers learn such topological structures of objects implicitly, enabling these controllers to achieve human-level manipulation?}

\section{Conclusion} \label{sec:conclusion}

We introduce a general learning-based garment-flattening policy \textit{LaGarNet} for single-gripper pick-and-place (\text{PnP}) action primitives. It successfully flattens all four garment types, including longsleeved T-shirt, trousers, skirts and dress, in both simulated and real-world environments.  We achieve this through a more generalised goal-conditioned recurrent state-space model, a data-collection strategy based on pre-trained diffusion policy, and a coverage-alignment reward function. \textit{LaGarNet} matches the state-of-the-art performance of mesh-based counterparts in garment flattening, marking the first successful application of state-space models on complex garments. 

\bibliographystyle{IEEEtran}
\bibliography{IEEEabrv,ref}

\end{document}